\documentclass[a4paper,hyperref]{rnti}

\usepackage[linesnumbered,ruled,vlined,norelsize]{algorithm2e}
 \usepackage{graphicx}

\usepackage[linesnumbered,ruled,vlined,norelsize]{algorithm2e}
\usepackage{threeparttable}
\usepackage{diagbox}
\usepackage{amsmath}

\usepackage{graphicx}
\usepackage{xcolor} 
\usepackage{dblfloatfix}

\usepackage[T1]{fontenc}
\usepackage[utf8]{inputenc}
\usepackage{mathabx}
\usepackage{wasysym}

\newcommand{\comment}[1]{}

\colorlet{myGreen}{green!40!gray}
\colorlet{myRed}{red}
\colorlet{myGray}{gray}

\usepackage{url}

\usepackage{adjustbox}

 \titrecourt{{\textquotedblleft}hasSignification(){\textquotedblright} : une nouvelle fonction pour soutenir la détection de DCP}
 \nomcourt {A. Mrabet et al.}

 \titre {%
{\textquotedblleft}hasSignification(){\textquotedblright} : une nouvelle fonction de distance pour soutenir la détection de données personnelles
}

 \auteur{Amine Mrabet, Ali Hassan, Patrice Darmon}
 \affiliation{%
    Research \& Innovation - Umanis\\
          7, Rue Paul Vaillant Couturier, 92300 Levallois-Perret, France\\
         \{ahassan, amrabet, pdarmon\} @umanis.com\\
 }

 \resume{
Avec le Big Data et les lacs de données d'aujourd'hui, nous sommes en face d'une masse de données difficile à gérer manuellement. La protection de données personnelles dans ce contexte nécessite donc une analyse automatique pour la découverte des données. Un stockage dans une base de connaissances des noms des attributs déjà analysés pourrait optimiser cette découverte automatique. Pour avoir une meilleure base de connaissances, il ne faut pas stocker les attributs dont le nom n'a pas de sens. 
Dans ce papier, pour vérifier si le nom d'un attribut a une signification, nous proposons de calculer les distances entre ce nom et l'ensemble des mots d'un dictionnaire. 
Nos études sur les fonctions de distance N-Gramme, Jaro-Winkler et Levenshtein montrent des limites pour fixer un seuil d'acceptation d'un attribut dans la base de connaissances. 
Afin de surmonter ces limites, notre solution vise à renforcer le calcul de score en utilisant une fonction exponentielle basée sur la séquence la plus longue. 
En outre, un double parcours de dictionnaire est également proposé afin de traiter les attributs qui ont un nom composé. 
}

 \summary{
Today with Big Data and data lakes, we are faced of a mass of data that is very difficult to manage it manually. The protection of personal data in this context requires an automatic analysis for data discovery. 
Storing the names of attributes already analyzed in a knowledge base could optimize this automatic discovery. To have a better knowledge base, we should not store any attributes whose name does not make sense.
In this article, to check if the name of an attribute has a meaning, we propose a solution that calculate the distances between this name and the words in a dictionary.
Our studies on the distance functions like N-Gram, Jaro-Winkler and Levenshtein show limits to set an acceptance threshold for an attribute in the knowledge base.
In order to overcome these limitations, our solution aims to strengthen the score calculation by using an exponential function based on the longest sequence.
In addition, a double scan in dictionary is also proposed in order to process the attributes which have a compound name.
}

 \usepackage{color}
 \definecolor{dark-blue}{rgb}{0,0,.35}
 \definecolor{dark-red}{rgb}{0.35,0,.35}
 \definecolor{dark-green}{rgb}{0,0.35,0}

 \hypersetup{
 filecolor=dark-green, 
 citecolor=dark-red, 
 colorlinks=true,
 }

\begin{document}

\section{Introduction}
L'évolution technologique des objets connectés et l'arrivée des techniques de Big Data, Data Lake et le Cloud ont augmenté considérablement la capacité de stockage et de collection des données. Ces dernières ont encouragé de plus en plus les entreprises à collecter des données externes qui viennent par exemple des réseaux sociaux et des applications sur les téléphones portables pour des objectifs commerciaux comme l'analyse de comportement des clients afin de proposer des nouveaux services ou produits adaptés. 
Ces données collectées contiennent souvent des données personnelles. Elles sont sous plusieurs formes et structures. 
Les entreprises qui reçoivent les données ne définissent pas forcément les métas qui décrivent ces données (comme le nom et le type des attributs collectés).
Par conséquent, il n'y a aucune garantie que les noms des attributs ont une signification qui aide les entreprises à comprendre et maîtriser les données. Par exemple, dans le contexte de Big Data où les données sont stockées sous une forme de {\textquotedblleft}\textit{clé:valeur}{\textquotedblright}, les collecteurs de données se retrouvent dans le choix de réduire la taille de la 'clé' en utilisant par exemple des abréviations. Ce qui permet de réduire le volume total de données en garantissant la qualité.

Donc, l'étape de découverte des données est indispensable pour les entreprises afin de maîtriser les données en général et de localiser les données personnelles, telles que les informations de santé et les données bancaires. 
Le but est de fournir des catalogues de données sensibles au sein d'une organisation.
La découverte des données est la première étape d'une élaboration réussite des stratégies de gouvernance des données par les propriétaires des données.

Une automatisation de cette étape est proposée dans \cite{MrabetBD19}] 
pour les bases de données structurées et semi-structurées et dans \cite{MrabetHD19}]
pour les bases de données multidimensionnelles. 
L'automatisation de cette étape permet d'éviter les étapes chronophages, d'augmenter la précision de la détection et essentiellement permet de garantir la confidentialité des données à caractère personnel (DCP) dans les différentes phases de collection, stockage, analyse et traitement des données \cite{Agarwal2020}.

La suite de ce papier est structurée de la façon suivante~: la section~\ref{sec:Motiv} montre le contexte et la motivation. 
La section~\ref{sec:Etatart} présente l'état de l'art.
Ensuite, la section~\ref{sec:casetude} montre des études préliminaires sur des fonctions de distance.
La section~\ref{sec:Proposition} détaille nos propositions pour renforcer le calcul de distance et traiter les noms des attributs composés. 
Nous terminons ce papier avec des expérimentations concernant la performance dans la section~\ref{sec:exp} et la conclusion en section~\ref{sec:Conclusion}.

\section{Contexte et motivation}
\label{sec:Motiv}
Afin d'optimiser la découverte automatique des DCP, 
il est proposé dans \cite{MrabetBD19,MrabetHD19}
d'utiliser une base de connaissances. L'idée principale de cette base est de stocker les connaissances collectées sur les attributs déjà analysés auparavant. Ensuite, pour chaque attribut à analyser, le système de la découverte de DCP exécute une recherche dans la base de connaissances afin de vérifier si cet attribut s'y trouve. Si cette base stocke assez de connaissances sur l'attribut (ç-a-d que l'attribut est déjà analysé plusieurs fois auparavant), on peut se baser sur ces connaissances pour décider si l'attribut est personnel ou pas sans effectuer des analyses aux niveaux données. Cette décision via la base de connaissances limite l'accès aux données ce qui améliore la confidentialité et réduit le temps d'analyse. 

La figure~\ref{fig:ontology} présente un extrait de la base de connaissances. Les informations stockées dans la base de connaissances concernent :
\begin{itemize}
    \item Le nom de l'attribut \textit{Attribut},
    \item Le contexte : \textit{Domaine} et \textit{Table},
    \item Les nombres des lignes analysées : \textit{Taille} et \textit{Taille\_Totale}
    \item Les pourcentages des lignes \textit{Taux} et \textit{Taux\_Final} qui correspondent à une DCP \textit{Référence}.
\end{itemize}

\begin{figure}[h]
\begin{scriptsize}
\begin{verbatim}
{"Domaine":"Vente",
 "Table":"Client",
 "Attribut":"Client_N",
 "Taille":[120,250],
 "Taille_Totale":370,
 "Scoring":[ {"Référence":"Nom personne","Taux":[60,70],"Taux_Final":67},
             {"Référence":"Prénom personne","Taux":[23,25],"Taux_Final":24},
             {"Référence":"Rue","Taux":[5,2],"Taux_Final":3}]},
		   
{"Domaine":"Vente",
 "Table":"Client",
 "Attribut":"Client_PN",
 "Taille":[120,250],
 "Taille_Totale":370,
 "Scoring":[ {"Référence":"Nom personne","Taux":[30,27],"Taux_Final":28},
             {"Référence":"Prénom personne","Taux":[80,77],"Taux_Final":78},
             {"Référence":"Rue","Taux":[2,0],"Taux_Final":1}]}
\end{verbatim}
\end{scriptsize}
	\caption{Extrait de la base de connaissances}
	\label{fig:ontology}
\end{figure}

Avec l'usage des abréviations qui augmente de plus en plus dans le contexte de Big Data, la base de connaissances risque de stocker des attributs dont le nom ne permet pas de déterminer si les données sont des DCPs ou non. Les abréviations pourraient avoir des usages multiples et différents d'une base de données à une autre. Donc, pour avoir une meilleure base de connaissances, il faut se limiter aux attributs dont le nom a un sens. Pour cette raison l'alimentation d'une base de connaissances propre nécessite une vérification sémantique (par rapport à des dictionnaires linguistiques) des noms des attributs à stocker. 

Pour faire cette vérification, nous proposons une fonction {\textquotedblleft}hasSignification(){\textquotedblright} qui calcule les distances entre les noms des attributs et l'ensemble de mots d'un dictionnaire de la langue utilisée dans la base de données. Le résultat de ce calcul est utilisé pour filtrer les attributs qui vont être stockés dans la base de connaissances. Si la fonction de distance trouve que le nom de l'attribut est très proche (à partir d'un seuil à choisir) d'un mot dans le dictionnaire, dans ce cas cet attribut doit être stocké dans la base de connaissances. 

\begin{figure}[h]
	\centering
	\includegraphics[width=1.0\columnwidth]{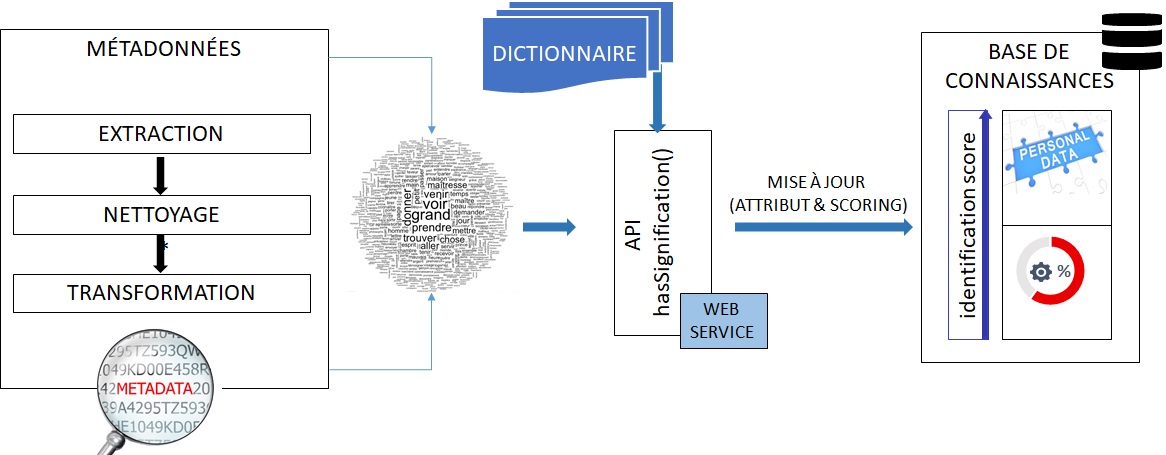}
	\caption{{\textquotedblleft}hasSignification(){\textquotedblright} et l'enrichissement de la base de connaissances}
	\label{fig:update_ontology}
\end{figure}

La figure~\ref{fig:update_ontology} montre que cette fonction {\textquotedblleft}hasSignification(){\textquotedblright} est utilisée dans les APIs de la base de connaissances pour filtrer les attributs. Après l'extraction des noms des attributs et avant d'appliquer le filtre sur ces noms, un nettoyage et une transformation sont nécessaires pour traiter les caractères spéciaux, les accents et les majuscules/minuscules.  

Pour calculer les distances entre le nom d'un attribut et les mots du dictionnaire, la fonction {\textquotedblleft}hasSignification(){\textquotedblright} devrait utiliser une fonction de distance. Notre objectif dans ce papier donc est de proposer une fonction de distance qui permet de sélectionner un seuil d'acceptation des attributs dans la base de connaissances. Cette fonction devrait être capable d'une manière stricte de distinguer entre les noms des attributs qui ont forcément une signification, ceux qui ont probablement une signification et ceux qui n'ont pas de signification.

\section{Positionnement et État de l'art}
\label{sec:Etatart}

Notre fonction {\textquotedblleft}hasSignification(){\textquotedblright} pourrait utiliser une fonction de distance qui existe dans la littérature. C'est pourquoi nous étudions les fonctions de distance les plus utilisées. 

Les fonctions de distances peuvent être classifiées selon leur approche dans plusieurs catégories \cite{Gomaa2013ASO} : 
\begin{itemize}
    \item \textbf{\textit{String-Based}} : le calcul de la similitude ou la distance entre deux chaînes fonctionne sur les séquences des caractères et la composition de chaînes.

    \item \textbf{\textit{Knowledge-Based}} : La similarité basée sur les connaissances est l'une des mesures de similarité sémantique entre les mots qui s'appuie sur les informations dérivées de réseaux sémantiques \cite{DBLP:conf/aaai/MihalceaCS06} comme par exemple {\textquotedblleft}WordNet{\textquotedblright} \cite{Miller1990IntroductionTW}.

    \item \textbf{\textit{Corpus-Based}} : le calcul de distance basé sur le corpus\footnotemark[1] est une comparaison qui détermine la distance sémantique entre les mots en fonction des informations obtenues à partir du corpus \cite{DBLP:conf/aaai/MihalceaCS06}. 
    \footnotetext[1]{Un corpus est une grande collection de textes écrits ou parlés qui est utilisé pour la recherche linguistique.}
    
    \item \textbf{\textit{Hybride}} : L'idée est de combiner les approches décrites précédemment pour atteindre un meilleur résultat en adoptant leurs avantages. 
    
\end{itemize}

\begin{table}[h]
\setlength{\tabcolsep}{3.4pt}
\begin{center}
\begin{tabular}{|c|cc|cc|cc|cc|}
\hline
\diagbox{ATT}{FD} & \multicolumn{2}{c|}{\textbf{2-Gramme}} & \multicolumn{2}{c|}{\textbf{3-Gramme}} & \multicolumn{2}{c|}{\textbf{Jaro-Winkler}} & \multicolumn{2}{c|}{\textbf{Levenshtein}} \\ 
\hline
 & \textbf{MPPD} & \textbf{SD} & \textbf{MPPD} & \textbf{SD} & \textbf{MPPD} & \textbf{SD} & \textbf{MPPD} & \textbf{SD} \\ 
\hline
{\color{myGreen}âgeclient} & client & 0.61 & client & 0.56 & âge & 0.84 & client & 0.67 \\ 
\hline
{\color{myGreen}naissance} & naissance & 1 & naissance & 1 & naissance & 1 & naissance & 1 \\ 
\hline
{\color{myGreen}médecintraitant} & traitant & 0.5 & médecin légiste & 0.53 & médecin & 0.91 & traitant & 0.53 \\ 
\hline
{\color{myGray}naiss} & naisse & 0.83 & naisse & 0.83 & naisse & 0.97 & naisse & 0.83 \\ 
\hline
{\color{myGray}dnaiss} & délaissé & 0.63 & divise & 0.58 & dais & 0.9 & naisse & 0.67 \\ 
\hline
{\color{myGray}adr-post} & aéroport & 0.63 & aéroport & 0.65 & adroit & 0.69 & aéroport & 0.63 \\ 
\hline
{\color{myRed}add pst} & adverse & 0.5 & addenda & 0.57 & adopte & 0.8 & budapest & 0.5 \\ 
\hline
{\color{myRed}comimuae} & commune & 0.69 & comiques & 0.75 & comique & 0.92 & commune & 0.75 \\ 
\hline
{\color{myRed}abteofkf} & attentif & 0.5 & attentif & 0.52 & abrite & 0.78 & attentif & 0.5 \\ 
\hline
{\color{myRed}abateofkf} & abattoir & 0.61 & abattoir & 0.66 & abattre & 0.85 & abattoir & 0.56 \\ 
\hline
\end{tabular}
\end{center}
\begin{tablenotes}
    \item FD : La Fonction de Distance
    \item MPPD : Le Mot le Plus Proche dans le Dictionnaire.
    \item SD : Le Score de Distance. 
    \item ATT : Le nom de l'attribut
\end{tablenotes}
\caption{Études préliminaires}
  \label{tab:data_études_préliminaires}
\end{table}

Les fonctions des catégories \textit{Knowledge-Based}, \textit{Corpus-Based} et \textit{Hybride} d'un côté nécessitent une phase importante de préparation pour qu'elles soient adaptées à plusieurs domaines à la fois. D'un autre côté, comme les travaux de \cite{IJAIN152} le montre, elles ne sont pas adaptées à calculer les distances pour des mots qui ne sont pas dans leurs bases de connaissances ou leurs corpus comme les abréviations que l'on pourrait les rencontrer potentiellement.

Ainsi, la catégorie \textit{String-Based} correspond le mieux à notre problématique du calcul de distance entre le nom d'un attribut et les mots du dictionnaire parce que la distance recherchée ce n'est pas une distance sémantique. 
La littérature est riche en propositions de fonctions de distance dans cette catégorie. Un état de l'art détaillé se trouve dans \cite{Gomaa2013ASO,IJAIN152}. 
Parmi les fonctions de distance les plus populaires nous citons les suivantes :
\begin{itemize}
\item 
\textbf{\textit{Longest Common Subsequence (LCS)}} : La distance entre deux chaînes est basée sur la longueur de la chaîne contiguë de caractères communs dans les deux chaînes \cite{Bergroth2000}.
\item 
\textbf{\textit{N-Gramme}} : N-gramme est une sous-séquence de \textit{N} éléments d'un texte donné. 
Les algorithmes de similarité comparent les N-grammes dans les deux chaînes de caractères. Des N-grammes supplémentaires pourraient être ajoutés pour renforcer la similitude des chaînes qui commencent et se terminent par les mêmes \textit{N} caractères. La distance est calculée en divisant le nombre de N-grammes similaires par le nombre de N-grammes uniques \cite{Kondrak2005,Hagree2019}. 
\item 
\textbf{\textit{Levenshtein}} : La distance de Levenshtein entre deux mots est le nombre minimal d'opérations (insertion, suppression ou substitution d'un seul caractère) requises pour transformer un mot en l'autre \cite{Levenshtein_SPD66}. Elle peut être normalisée en la divisant par la longueur du mot le plus long \cite{Yujian2007}.
\item 
\textbf{\textit{Jaro}} est basé sur le nombre et l'ordre des caractères communs entre deux chaînes \cite{Jaro1989}. Le résultat de cette distance est entre zéro (les deux chaînes sont complètement différentes) et un (les deux chaînes sont identiques). 
\item 
\textit{\textbf{Jaro-Winkler}} est une extension de la distance Jaro qui donne des évaluations plus importantes pour les chaînes qui ont le même préfixe \cite{winkler90}. 
\end{itemize}

\section{Études préliminaires}
\label{sec:casetude}

Nos travaux commencent par des études préliminaires sur les trois fonctions de distance de la catégorie \textit{String-Based} les plus utilisées~: N-Gramme, Levenshtein et Jaro-Winkler. Concernant N-Gramme, nos études se concentrent sur 2-Gramme et 3-Gramme. 
L'objectif de ces travaux est de vérifier la possibilité d'utiliser une de ces fonctions pour répondre à nos besoins pour l'enrichissement de la base de connaissances.

Ainsi dans ces études, la fonction {\textquotedblleft}hasSignification(){\textquotedblright} utilise chacune de ces quatre fonctions (2-Gramme, 3-Gramme, Levenshtein et Jaro-Winkler) pour calculer les distances entre les noms des attributs et l'ensemble de mots d'un dictionnaire.
Par ailleurs, le résultat de la fonction {\textquotedblleft}hasSignification(){\textquotedblright} est normalisé pour qu'il représente la probabilité que le mot concerné (le nom de l'attribut) ait une signification. Autrement dit, le résultat de la fonction {\textquotedblleft}hasSignification(){\textquotedblright} est entre {\textquotedblleft}0{\textquotedblright} (cela veut dire que le mot n'a pas de signification) et {\textquotedblleft}1{\textquotedblright} (cela veut dire que le mot a forcement une signification).

Pour effectuer ces études, nous utilisons un ensemble de noms des attributs potentiels classifiés dans trois catégories :
\begin{itemize}
    \item Les noms des attributs (mots) qui ont forcément une signification~: Chaque nom correspond à un mot du dictionnaire ou plusieurs concaténés. Cette catégorie est présenté en utilisant la couleur verte dans la suite du papier. Les mots qui représentent cette catégorie dans nos études sont {\textquotedblleft}âgeclient{\textquotedblright}, {\textquotedblleft}naissance{\textquotedblright} et {\textquotedblleft}médecintraitant{\textquotedblright}.
    
    \item Les noms des attributs qui ont probablement une signification~: Chaque nom correspond à une abréviation assez proche du mot d'origine. Cette catégorie est présentée en gris. Les mots qui représentent cette catégorie sont {\textquotedblleft}naiss{\textquotedblright}, {\textquotedblleft}dnaiss{\textquotedblright} et {\textquotedblleft}adr-post{\textquotedblright} qui sont des abréviations des mots {\textquotedblleft}naissance{\textquotedblright}, {\textquotedblleft}date de naissance{\textquotedblright} et {\textquotedblleft}adresse postale{\textquotedblright} respectivement.
    
    \item Les noms des attributs qui n'ont pas de signification : Chaque nom correspond à une abréviation très loin du mot d'origine ou à une chaîne de caractères aléatoire. Cette catégorie est présentée en rouge. Les mots qui appartiennent à cette catégorie sont {\textquotedblleft}add pst{\textquotedblright}, {\textquotedblleft}comimuae{\textquotedblright}, {\textquotedblleft}abteofkf{\textquotedblright} et {\textquotedblleft}abateofkf{\textquotedblright}
\end{itemize}

La table~\ref{tab:data_études_préliminaires} présente les résultats de l'implémentation de la fonction {\textquotedblleft}hasSignification(){\textquotedblright} avec chacune des quatre fonctions de distance de la littérature sur la liste d'attributs potentiels. Les colonnes {\textquotedblleft}MPPD{\textquotedblright} contiennent les mots de dictionnaire les plus proches aux noms des attributs selon chaque fonction de distance. Les colonnes {\textquotedblleft}SD{\textquotedblright} contiennent les scores de distance normalisés entre les mots les plus proches et les noms des attributs. 

Afin de sélectionner un seuil d'acceptation pour la base de connaissances, la fonction de distance devrait distinguer entre les trois catégories de noms des attributs (verte, grise et rouge). Autrement dit, tous les mots de la catégorie verte devraient avoir des scores plus élevés que tous les mots de la catégorie grise qui sont à leur tour devraient avoir des scores plus élevés que tous les mots de la catégorie rouge. La table~\ref{tab:data_études_préliminaires} montre qu'aucune de quatre fonctions distingue entre les trois catégories. Par exemple, pour les quatre fonctions, le mot {\textquotedblleft}comimuae{\textquotedblright} de la catégorie rouge a un score plus élevé que le mot {\textquotedblleft}adr-post{\textquotedblright} de la catégorie grise et même plus élevé que le mot {\textquotedblleft}médecintraitant{\textquotedblright} de la catégorie verte. Pour cette raison, nous proposons dans la section suivante une amélioration du calcul de distance entre les noms des attributs et l'ensemble de mots d'un dictionnaire. 

\section{La nouvelle {\textquotedblleft}hasSignification(){\textquotedblright}}
\label{sec:Proposition}
Afin d'améliorer le calcul de distance obtenu dans la section~\ref{sec:casetude} de la fonction {\textquotedblleft}hasSignification(){\textquotedblright} proposée dans la section~\ref{sec:Motiv}, nous redéfinissons l'algorithme de cette fonction en se basant sur deux étapes~:
\begin{enumerate}
    \item Un premier parcours de dictionnaire pour traiter les noms des attributs composés,
    \item Un deuxième parcours pour calculer les distances entre les noms des attributs et les mots du dictionnaire. Ce calcul est renforcé par une fonction exponentielle dont l'exposant est basé sur la séquence commune la plus longue \cite{Bergroth2000}. 
\end{enumerate}

L'algorithme~\ref{algo:hasSematic} détaille nos propositions. Il prend en entrée le dictionnaire linguistique et la liste des noms des attributs dont il faut vérifier la signification. 
Le dictionnaire doit être nettoyé des prépositions, des pronoms, des articles définis et indéfinis, etc.
L'algorithme commence par une étape d'ordonnancement de mots du dictionnaire selon leur taille à la place de l'ordre alphabétique (ligne 1). Cet ordonnancement est nécessaire pour traiter les noms des attributs composés. 

Pour chaque attribut, dans le premier parcours (ligne 4 à 8), l'algorithme vérifie pour chaque mot de dictionnaire ayant une taille supérieure à trois (ligne 4) si le nom de l'attribut contient ce mot (ligne 6). Si c'est le cas, le mot va être enlevé du nom de l'attribut (ligne 7) et la taille de ce mot va être prise en compte (ligne 8) pour renforcer le calcul de distance dans le deuxième parcours. 

L'ordonnancement de dictionnaire selon la taille de mots permet d'éviter d'enlever une partie d'un mot long qui correspond à un mot plus court qui se trouve avant dans le dictionnaire. Par exemple, dans un dictionnaire ordonné alphabétiquement, le mot {\textquotedblleft}trait{\textquotedblright} se trouve avant le mot {\textquotedblleft}traitant{\textquotedblright}. En respectant cet ordre, si le nom de l'attribut inclut le mot {\textquotedblleft}traitant{\textquotedblright}, la partie {\textquotedblleft}trait{\textquotedblright} va être enlevée en gardant le reste du mot {\textquotedblleft}ant{\textquotedblright}. Ce qui impact considérablement le calcul de distance. En outre, dans ce premier parcours, les mots du dictionnaire courts (ayant une taille inférieure à quatre) ne sont pas pris en compte. Ce choix est fait parce que même une chaîne de caractères aléatoire peut inclure un mot court.

\begin{algorithm}[!h]
%
	\caption{{\sc } hasSignification()}
	\label{algo:hasSematic}

	\KwIn{$Dictionnaire \{mot_k\}, ATT_{name}: $ liste des attributs}
	\KwOut{$result \langle att_i , SD\_MPPD_i \rangle$}
	$Dictionnaire.sortBy(mot_i.length)$\;
	\ForEach{$att_i \in ATT_{name} $}
	{ 
	    $s \gets 0$\;
	    \ForEach{$mot_i \in Dictionnaire |mot_i.length>3$}{
	        $APP \gets att_i$\;
			\If{($APP.contain(mot_i)$)} 
        	    {$APP \gets app.remplace (mot_i,$ {\textquotedblleft}{\textquotedblright}$)$\;
        	    $s \gets s+mot_i.length$\;}
        	    } 
       	\If{($APP$ is Empty) $\lor$ ($APP$ isnot Alpha)}
        	   {$result.put(att_i, 1)$\;}	  
        \Else{
            $SD \gets new Map\langle double, string \rangle$\;
			\ForEach{$mot_j \in Dictionnaire $}{
			   $SD.put(FDistance(APP,mot_j), mot_j)$\;
            }
            $SD\_MPPD_i \gets Max(SD.keys)$\;
            $MPPD_i \gets SD.get(SD\_MPPD_i)$\;
            $lcs \gets LCS(APP,MPPD_i).length$\;
            $s \gets s + lcs$\;
            $\overline{s} \gets APP.length + MPPD_i.length - 2\times lcs$\;
                $SD\_MPPD_i \gets SD\_MPPD_i^{\overline{s}/s}$\;
                $result.put(att_i, SD\_MPPD_i)$\;}
    }
    {\Return{result}\;}    
\end{algorithm}

À la fin de ce premier parcours, s'il ne reste aucune lettre dans le nom de l'attribut (ligne 9) alors l'algorithme considère que ce nom a une signification et il l'associe à un score {\textquotedblleft}1{\textquotedblright} (ligne 10). 
Si ce n'est pas le cas, un deuxième parcours est nécessaire pour calculer la distance normalisée entre le reste du nom de l'attribut après le premier parcours ({\textquotedblleft}\textit{APP}{\textquotedblright}) et chaque mot du dictionnaire (lignes 12 à 14). 
La fonction {\textquotedblleft}\textit{FDistance()}{\textquotedblright} de la ligne 14 qui calcule cette distance représente une des quatre fonctions étudiées dans la section~\ref{sec:casetude}. 
Les scores de distance sont stockés dans une variable {\textquotedblleft}\textit{SD}{\textquotedblright} de type {\textquotedblleft}\textit{clé:valeur}{\textquotedblright} qui associe les scores (clés) obtenus aux mots du dictionnaire (valeurs). 
Après ce deuxième parcours, l'algorithme ensuite trouve le score le plus élevé ({\textquotedblleft}\textit{$SD\_MPPD_i$}{\textquotedblright} de la ligne 15) et le mot du dictionnaire correspondant ({\textquotedblleft}\textit{$MPPD_i$}{\textquotedblright} de la ligne 16). Ce mot est le plus proche de {\textquotedblleft}\textit{APP}{\textquotedblright}.

\begin{table}[h]
\setlength{\tabcolsep}{3.4pt}
\begin{center}
\begin{tabular}{|c|c|cc|cc|cc|cc|}
\hline
\diagbox{ATT}{FD} & \multicolumn{1}{c|}{APP} & \multicolumn{2}{c|}{2-Gramme} & \multicolumn{2}{c|}{3-Gramme} & \multicolumn{2}{c|}{Jaro-Winkler} & \multicolumn{2}{c|}{Levenshtein} \\ 
\hline
& & MPPD & SD & MPPD & SD & MPPD & SD & MPPD & SD \\ 
\hline
{\color{myGreen}âgeclient} & âge & âge & 1 & âge & 1 & âge & 1 & âge & 1 \\ 
\hline
{\color{myGreen}naissance} & $\emptyset$ & $\emptyset$ & 1 & $\emptyset$ & 1 & $\emptyset$ & 1 & $\emptyset$ & 1 \\ 
\hline
{\color{myGreen}médecintraitant} & $\emptyset$ & $\emptyset$ & 1 & $\emptyset$ & 1 & $\emptyset$ & 1 & $\emptyset$ & 1 \\ 
\hline
{\color{myGray}naiss} & // & naisse & 0.96 & naisse & 0.96 & naisse & 0.99 & naisse & 0.96 \\ 
\hline
{\color{myGray}dnaiss} & // & délaissé & 0.49 & divise & 0.12 & dais & 0.87 & naisse & 0.85 \\ 
\hline
{\color{myGray}adr-post} & adr- & aura & 0.57 & adam & 0.83 & adroit & 0.9 & madre & 0.8 \\ 
\hline
{\color{myRed}add pst} & // & adverse & 0.03 & addenda & 0.22 & adopte & 0.37 & budapest & 0.02 \\ 
\hline
{\color{myRed}comimuae} & // & commune & 0.32 & comiques & 0.56 & comique & 0.86 & commune & 0.42 \\ 
\hline
{\color{myRed}abteofkf} & // & attentif & 0.02 & attentif & 0.02 & abrite & 0.29 & attentif & 0.02 \\ 
\hline
{\color{myRed}abateofkf} & // & abattoir & 0.33 & abattoir & 0.4 & abattre & 0.72 & abattoir & 0.27 \\ 
\hline
\end{tabular}

\end{center}

\begin{tablenotes}
    \item APP : le nom de l'attribut qui reste Après le Premier Parcours.
\end{tablenotes}
\caption{Extrait des données\\}
\label{tab:data}
\end{table}

Ce score va être renforcé dans la suite de l'algorithme en appliquant une fonction exponentielle. L'exposant de cette fonction est un quotient de ($\overline{s}$) sur ($s$) (ligne 20). Où : 
\begin{itemize}
    \item ($s$) cumule (ligne~18) :
    \begin{itemize}
        \item Les tailles des mots enlevés dans le premier parcours (ligne 8) avec
        \item La taille de la plus longue séquence commune (lcs) entre {\textquotedblleft}\textit{APP}{\textquotedblright} et {\textquotedblleft}\textit{$MPPD_i$}{\textquotedblright} (ligne~17).
    \end{itemize}
    \item ($\overline{s}$) est le nombre de caractères de {\textquotedblleft}\textit{$MPPD_i$}{\textquotedblright} et {\textquotedblleft}\textit{APP}{\textquotedblright} qui n'appartiennent pas à lcs (ligne 19).
\end{itemize}
Enfin, l'algorithme associe le nouveau score au nom d'attribut concerné (ligne 21). 

La table~\ref{tab:data} présente les résultats de l'application de cet algorithme sur l'ensemble des noms des attributs présentés dans la section~\ref{sec:casetude} en utilisant chacune des quatre fonctions proposées dans la même section. 
Ces résultats montrent une amélioration variée pour les quatre fonctions. Grâce au premier parcours du dictionnaire, les quatre fonctions donnent un score {\textquotedblleft}1{\textquotedblright} à tous les noms des attributs de la catégorie verte ce qui permet bien évidemment de la distinguer des catégories grise et rouge. 

Bien que l'amélioration concerne les quatre fonctions, nous remarquons que les deux fonctions 3-Gramme et Jaro-Winkler ne permettent pas de distinguer entre les catégories rouge et grise. 
Par exemple, le score de {\textquotedblleft}comimuae{\textquotedblright} de la catégorie rouge est très proche et supérieur à celui de {\textquotedblleft}dnaiss{\textquotedblright} de la catégorie grise pour respectivement les fonctions Jaro-Winkler et 3-Gramme. 
Contrairement à ces deux dernières fonctions, les deux autres fonctions Levenshtein et 2-Gramme distinguent facilement les trois catégories. 

Nous rendons le code de notre fonction {\textquotedblleft}hasSignification(){\textquotedblright} open source et disponible sur ce lien \href{https://github.com/aloshi1981/hasSignification}{https://github.com/aloshi1981/hasSignification}

Des expérimentations sont détaillées dans la section suivante afin de choisir une de ces deux fonctions en se basant sur leur performance en terme de temps d'exécution.

\section{Expérimentations}
\label{sec:exp}
Dans cette section nous présentons deux expérimentations afin d'analyser le temps d'exécution de {\textquotedblleft}hasSignification(){\textquotedblright} selon :
\begin{itemize}
    \item la fonction de distance,
    \item la taille du dictionnaire.
\end{itemize} 
Les deux expérimentations sont réalisées sur une machine ayant un processeur Intel i5-8350U et 8 Go de RAM.

\subsection{Analyse de performance de {\textquotedblleft}hasSignification(){\textquotedblright} selon la fonction de distance}
Les études réalisées dans la section~\ref{sec:Proposition} montrent que l'utilisation des deux fonctions Levenshtein et 2-Gramme permet à {\textquotedblleft}hasSignification(){\textquotedblright} de répondre à nos besoins. Donc, dans cette expérimentation nous réalisons une analyse de performance pour choisir une.

\textbf{Protocole :} Nous avons réalisé deux implémentations de la fonction {\textquotedblleft}hasSignification(){\textquotedblright} une avec Levenshtein et l'autre avec 2-Gramme.
Plusieurs exécutions sont faites sur plusieurs groupes de 100, 1000, 2000, 5000 et 10000 noms d'attributs potentiels. Nous observons le temps d'exécution des deux implémentations pour chaque groupe.

\begin{figure}[h]
	\centering
	\includegraphics[width=0.9\columnwidth]{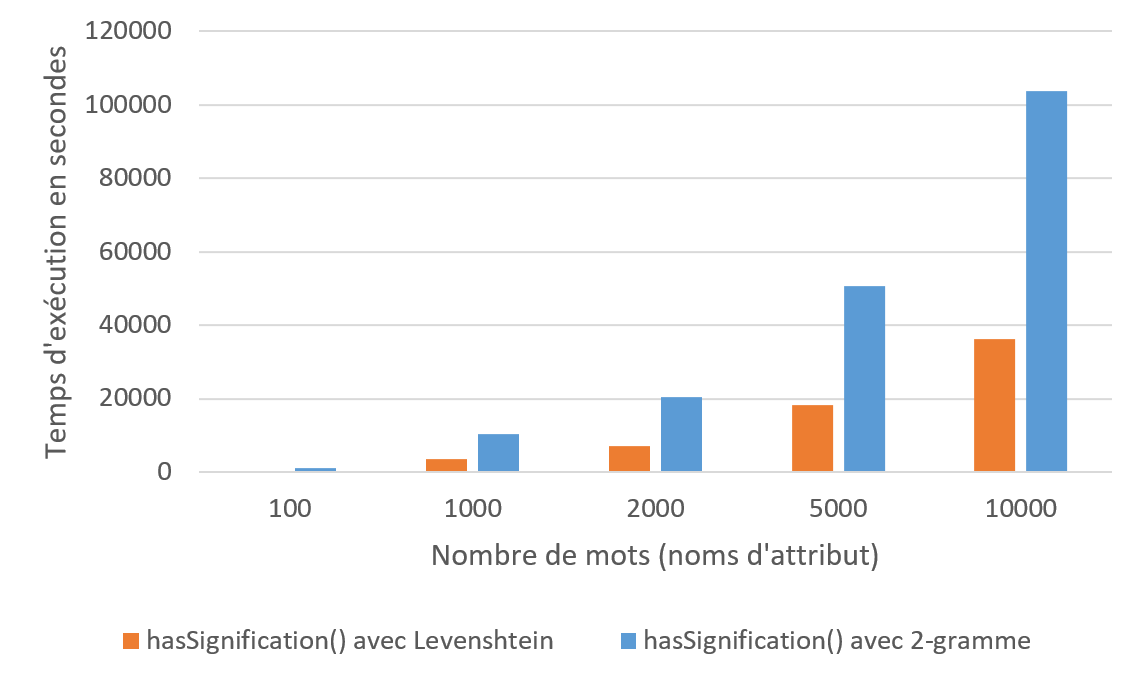}
	\caption{Analyse de temps d'exécution de {\textquotedblleft}hasSignification(){\textquotedblright} selon la fonction de distance}
	\label{fig:Analyse_fonction}
\end{figure}

\textbf{Résultats :}
La figure~\ref{fig:Analyse_fonction} montre les temps d'exécution des deux implémentations. Selon ces résultats, le temps d'exécution de {\textquotedblleft}hasSignification(){\textquotedblright} avec Levenshtein est largement inférieur à celui avec 2-Gramme.

En se basant sur ces résultats nous avons choisi d'utiliser {\textquotedblleft}hasSignification(){\textquotedblright} avec Levenshtein pour vérifier la signification des noms d'attributs.

\subsection{Analyse de performance de {\textquotedblleft}hasSignification(){\textquotedblright} selon la taille du dictionnaire}
Cette expérimentation étudie la relation entre le temps d'exécution de {\textquotedblleft}hasSignification(){\textquotedblright}, qui utilise la fonction de distance Levenshtein, et le nombre de mots dans le dictionnaire linguistique.

\textbf{Protocole :}
Nous utilisons pour cette deuxième expérimentation, deux versions du dictionnaire linguistique : une avec environ 22000 mots et l'autre avec environ 63000 mots.
De la même manière que la première expérimentation, nous avons fait plusieurs exécutions de {\textquotedblleft}hasSignification(){\textquotedblright} sur plusieurs groupes de 100, 1000, 2000, 5000 et 10000 noms d'attributs potentiels. Nous observons également le temps d'exécution de {\textquotedblleft}hasSignification(){\textquotedblright} en utilisant chaque version du dictionnaire pour chaque groupe.

\begin{figure}[h]
	\centering
	\includegraphics[width=0.9\columnwidth]{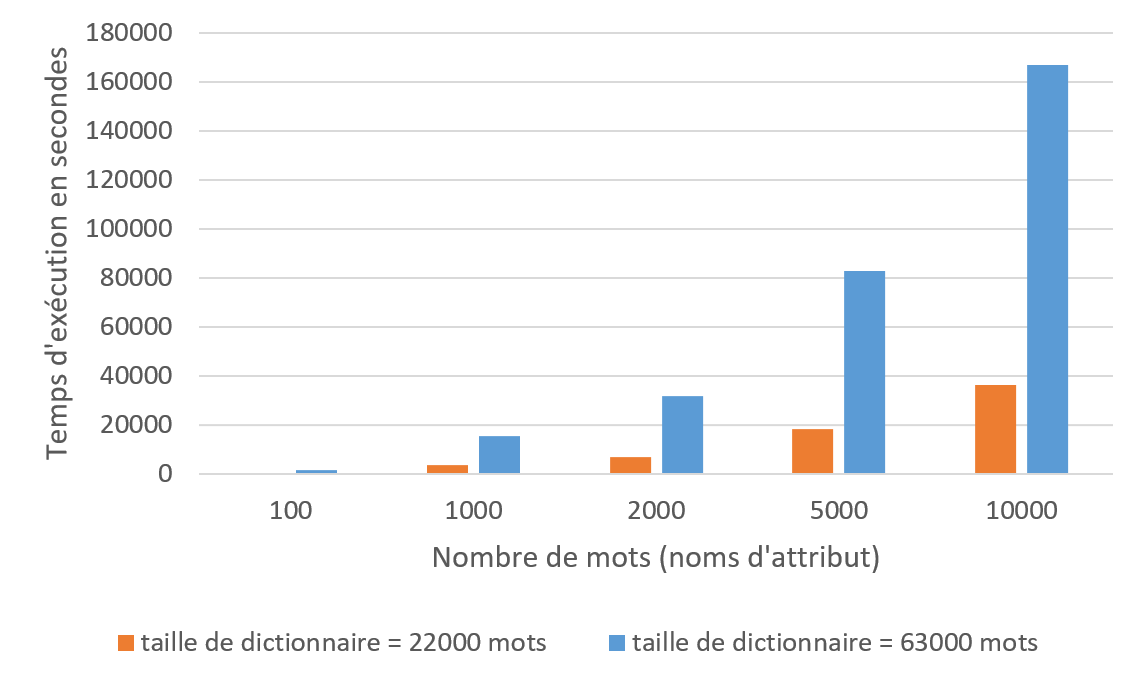}
	\caption{Analyse de temps d'exécution de {\textquotedblleft}hasSignification(){\textquotedblright} en fonction de la taille du dictionnaire}
	\label{fig:Analyse_Dic}
\end{figure}

\textbf{Résultats :}
La figure~\ref{fig:Analyse_Dic} montre les temps d'exécution correspondant aux deux versions du dictionnaire. Nous pouvons remarquer que le temps requis pour exécuter {\textquotedblleft}hasSignification(){\textquotedblright} avec un dictionnaire de 63000 mots est quatre fois supérieur au celui avec un dictionnaire de 22000. Ce qui montre l'importance d'utiliser un dictionnaire qui ne contient pas de mots inutiles comme les verbes conjugués.

\section{Conclusion}
\label{sec:Conclusion}
Nous avons présenté dans ce papier la fonction {\textquotedblleft}hasSignification(){\textquotedblright} qui calcule les distances entre des noms des attributs et l'ensemble de mots d'un dictionnaire afin de vérifier si ces noms ont des significations ou non. 
Cette fonction a pour but d'améliorer la qualité d'une base de connaissances utilisée pour optimiser la détection automatique des DCP. 
L'objectif est d'avoir une base de connaissances la plus propre possible, en stockant que les attributs dont le nom a une signification.

Les études sur les fonctions les plus utilisées dans la littérature N-Gramme, Jaro-Winkler et Levenshtein montrent des limites pour distinguer entre les noms des attributs qui ont forcément une signification, ceux qui ont probablement une signification et ceux qui n'ont pas de signification. C'est pourquoi la fonction {\textquotedblleft}hasSignification(){\textquotedblright} proposée se base sur un double parcours du dictionnaire. Le premier a pour objectif de traiter les noms des attributs composés.
Tandis que, le rôle du deuxième parcours est de calculer les distances entre le reste du nom de l'attribut après le premier parcours et les différents mots du dictionnaire.
Ce calcul est renforcé par une fonction exponentielle basée sur la taille de la plus longue séquence commune entre le nom de l'attribut et le mot du dictionnaire.

Les résultats de cette amélioration du calcul montrent que les deux fonctions 2-Gramme et Levenshtein répondent parfaitement à nos besoins. Par ailleurs, les études expérimentales réalisées montrent que le temps d'exécution de {\textquotedblleft}hasSignification(){\textquotedblright} avec Levenshtein est largement inférieur à celui avec 2-Gramme. C'est pourquoi nous avons choisi d'utiliser {\textquotedblleft}hasSignification(){\textquotedblright} avec Levenshtein afin de vérifier les significations des noms des attributs avant de les intégrer dans la base de connaissances.

{
\bibliographystyle{rnti}
\bibliography{references}

\providecommand\Fr{}
\providecommand\Eng{}
\providecommand\andname{and}
\providecommand\andnamec{and}

\begin{thebibliography}{}


\bibitem[{Agarwal et~al.}(2019){Agarwal, Gupta, \andnamec{}
  Sharma}]{Agarwal2020}
Agarwal, S., M.~Gupta, \andname{} A.~Sharma (2019).
\newblock {Big Data Privacy Issues Solutions}.
\newblock {\em Proceedings of the IEEE International Conference Image
  Information Processing\/}~{\em 2019-November}, 225--228.

\bibitem[{Al-Hagree et~al.}(2019){Al-Hagree, Al-Sanabani, Hadwan, \andnamec{}
  Al-Hagery}]{Hagree2019}
Al-Hagree, S., M.~Al-Sanabani, M.~Hadwan, \andname{} M.~A. Al-Hagery (2019).
\newblock An improved n-gram distance for names matching.
\newblock In {\em 2019 First International Conference of Intelligent Computing
  and Engineering (ICOICE)}, pp.\  1--7.

\bibitem[{Bergroth et~al.}(2000){Bergroth, Hakonen, \andnamec{}
  Raita}]{Bergroth2000}
Bergroth, L., H.~Hakonen, \andname{} T.~Raita (2000).
\newblock A survey of longest common subsequence algorithms.
\newblock In {\em Proceedings Seventh International Symposium on String
  Processing and Information Retrieval. SPIRE 2000}, pp.\  39--48.

\bibitem[{Gomaa \andnamec{} Fahmy}(2013){Gomaa \andnamec{}
  Fahmy}]{Gomaa2013ASO}
Gomaa, W.~H. \andname{} A.~Fahmy (2013).
\newblock A survey of text similarity approaches.
\newblock {\em International Journal of Computer Applications\/}~{\em 68},
  13--18.

\bibitem[{Jaro}(1989){Jaro}]{Jaro1989}
Jaro, M.~A. (1989).
\newblock Advances in record-linkage methodology as applied to matching the
  1985 census of tampa, florida.
\newblock {\em Journal of the American Statistical Association\/}~{\em
  84\/}(406), 414--420.

\bibitem[{Kondrak}(2005){Kondrak}]{Kondrak2005}
Kondrak, G. (2005).
\newblock N-gram similarity and distance.
\newblock In M.~Consens \andname{} G.~Navarro (Eds.), {\em String Processing
  and Information Retrieval}, Berlin, Heidelberg, pp.\  115--126. Springer
  Berlin Heidelberg.

\bibitem[{Levenshtein}(1966){Levenshtein}]{Levenshtein_SPD66}
Levenshtein, V.~I. (1966).
\newblock Binary codes capable of correcting deletions, insertions and
  reversals.
\newblock {\em Soviet Physics Doklady\/}~{\em 10\/}(8), 707--710.
\newblock Doklady Akademii Nauk SSSR, V163 No4 845-848 1965.

\bibitem[{Mihalcea et~al.}(2006){Mihalcea, Corley, \andnamec{}
  Strapparava}]{DBLP:conf/aaai/MihalceaCS06}
Mihalcea, R., C.~D. Corley, \andname{} C.~Strapparava (2006).
\newblock Corpus-based and knowledge-based measures of text semantic
  similarity.
\newblock In {\em Proceedings, The Twenty-First National Conference on
  Artificial Intelligence and the Eighteenth Innovative Applications of
  Artificial Intelligence Conference, {USA}}, pp.\  775--780.

\bibitem[{Miller et~al.}(1990){Miller, Beckwith, Fellbaum, Gross, \andnamec{}
  Miller}]{Miller1990IntroductionTW}
Miller, G., R.~Beckwith, C.~Fellbaum, D.~Gross, \andname{} K.~J. Miller (1990).
\newblock Introduction to wordnet: An on-line lexical database.
\newblock {\em International Journal of Lexicography\/}~{\em 3}, 235--244.

\bibitem[{{MRABET} et~al.}(2019){{MRABET}, {BENTOUNSI}, \andnamec{}
  {DARMON}}]{MrabetBD19}
{MRABET}, A., M.~{BENTOUNSI}, \andname{} P.~{DARMON} (2019).
\newblock Secp2i a secure multi-party discovery of personally identifiable
  information (pii) in structured and semi-structured datasets.
\newblock In {\em IEEE Intl. Conf. on Big Data (Big Data)}, pp.\  5028--5033.

\bibitem[{Mrabet et~al.}(2019){Mrabet, Hassan, \andnamec{} Darmon}]{MrabetHD19}
Mrabet, A., A.~Hassan, \andname{} P.~Darmon (2019).
\newblock D{\'{e}}tection des donn{\'{e}}es {\`{a}} caract{\`{e}}re personnel
  dans les bases multidimensionnelles.
\newblock In {\em Business Intelligence {\&} Big Data, 15{\`{e}}me Edition de
  la conf{\'{e}}rence EDA}, Volume {B-15} of {\em {RNTI}}, pp.\  31--44.

\bibitem[{Prasetya et~al.}(2018){Prasetya, Wibawa, \andnamec{}
  Hirashima}]{IJAIN152}
Prasetya, D.~D., A.~P. Wibawa, \andname{} T.~Hirashima (2018).
\newblock The performance of text similarity algorithms.
\newblock {\em International Journal of Advances in Intelligent
  Informatics\/}~{\em 4\/}(1), 63--69.

\bibitem[{Winkler}(1990){Winkler}]{winkler90}
Winkler, W.~E. (1990).
\newblock String comparator metrics and enhanced decision rules in the
  fellegi-sunter model of record linkage.
\newblock In {\em Proceedings of the Section on Survey Research}, pp.\
  354--359.

\bibitem[{Yujian \andnamec{} Bo}(2007){Yujian \andnamec{} Bo}]{Yujian2007}
Yujian, L. \andname{} L.~Bo (2007).
\newblock A normalized levenshtein distance metric.
\newblock {\em IEEE Transactions on Pattern Analysis and Machine
  Intelligence\/}~{\em 29\/}(6), 1091--1095.

\end{thebibliography}
}

\end{document}